# PSDNet: Determination of Particle Size Distributions Using Synthetic Soil Images and Convolutional Neural Networks

Javad Manashti[1, 5], Pouyan Pirnia[1], Alireza Manashty[3], Sahar Ujan[4], Matthew Toews[2], François Duhaime[1]

[1] *Laboratory for Geotechnical and Geoenvironmental Engineering (LG2), École de technologie supérieure*

[2] *Systems Engineering Department, École de technologie supérieure, Montreal, Canada*

[3] *Computer Science Department, University of Regina, Regina, Saskatchewan*

[4] *Electrical Engineering Department, École de technologie supérieure, Montreal, Canada*

[5] *Corresponding author, Laboratory for Geotechnical and Geoenvironmental Engineering (LG2), École de technologie supérieure, 1100 Notre-Dame Ouest, Montreal, Quebec, H3C 1K3, Canada*

**Abstract**: This project aimed to determine the grain size distribution of granular materials from images using convolutional neural networks. The application of ConvNet and pretrained ConvNet models, including AlexNet, SqueezeNet, GoogLeNet, InceptionV3, DenseNet201, MobileNetV2, ResNet18, ResNet50, ResNet101, Xception, InceptionResNetV2, ShuffleNet, and NASNetMobile was studied. Synthetic images of granular materials created with the discrete element code YADE were used. All the models were trained and verified with grayscale and color band datasets with image sizes ranging from 32 to 160 pixels. The proposed ConvNet model predicts the percentages of mass retained on the finest sieve, coarsest sieve, and all sieves with root-mean-square errors of 1.8 %, 3.3 %, and 2.8 %, respectively, and a coefficient of determination of 0.99. For pretrained networks, root-mean-square errors of 2.4 % and 2.8 % were obtained for the finest sieve with feature extraction and transfer learning models, respectively.

# 1. Introduction

The particle size distribution (PSD) describes the statistical distribution of particle sizes in granular materials. Sieving [1] is the standard method to determine the PSD of granular materials. The PSD characterizes the cumulative mass percentages passing a series of sieves with different opening sizes. Sieving is accurate, but it is time-consuming and difficult to adapt to online processing.

PSD determination is of great importance in most geotechnical engineering projects. For instance, materials used in construction projects such as embankment dams or roads are selected based on their PSD. Many soil properties, such as permeability, can be estimated based on the material PSD. The PSD also plays a primary role in the chemical and metallurgical industries [2]. Ore processing has PSD requirements [3, 4], and online PSD determination is fundamental for performance improvement [5]. It is also critical to detect and determine particle size distributions in the food [6], pharmaceutical [7], and chemical industries [8].

Image-based methods have been used for PSD determinations for several decades. These methods provide a fast, non-contact, and economical way to determine the PSD, but they are not without disadvantages. Thurley and Ng [9] divided the sources of error into segregation and grouping error, profile error, capturing error, and overlapping-particle error. These errors are caused by the photographs only capturing the surface of the material. Photographs are in effect "sampling" the material, and are unable to describe the whole sample.

Image analysis techniques can be divided into two groups: direct and indirect methods. Direct methods use image segmentation for PSD determination, while indirect methods extract textural features from the images which are then used to determine the PSD.

Most commercial codes use direct methods to determine the PSD. Various algorithms are used to detect the contours of each particle in the image. The PSD is estimated using the size of the segmented area and statistical principles. The accuracy of direct methods was appraised by Liu and Tran [10] by comparing the $D_{50}$ obtained through sieving and a series of commercial software packages for mine backfill material. The $D_{50}$ is the particle size for which 50 % of the mass if composed of smaller particles. Compared to sieving, the error on $D_{50}$ was between 212 and 224 % for FragScan [11], 32 and 60 % for WipFrag [12], and between 37 and 48 % for Split [13]. The results presented by Liu and Tran [10] can also be used to compute the root-mean-square error (

RMSE) on the percentages passing for each sieve. For these three commercial codes, the RMSE varied between 13 and 36 %. Codes for particle size analysis in other fields, for instance BASEGRAIN in hydraulics and sediment transport, are also often based on segmentation [14].

The textural features used with indirect methods can be divided into three groups: statistical, local pattern, and transform-based features [15]. Haralick features [16], local entropy [17–19], and histogram of oriented gradients (HOG) [19, 20] are based on statistics of pixel intensity. Features based on local patterns establish a relationship between the gray level of each pixel in an image and the pixels located in their neighborhood [21]. Local binary pattern (LBP) [21], local configuration pattern (LCP) [22], and completed local binary pattern (CLBP) [23] are examples of local pattern features. Transform-based features describe the image texture in the frequency domain [24]. These features include Haar wavelet transforms [19, 25, 26], Fourier transforms [26, 27], and Gabor filters [26, 28].

The textural features that are most commonly used for PSD determination in geotechnical engineering are Haar wavelet transforms. Hryciw et al. [25] presented the theoretical basis for PSD determination with Haar wavelet transforms. These features form the basis of the Sedimaging system described by Ohm and Hryciw [29]. This system combines mechanical grain size sorting through sedimentation in a water column with textural analysis of photographs of the material after sedimentation. Based on the PSD presented by Ohm and Hryciw for clean sand [29], a RMSE on the percentages passing of 7.4 % can be calculated when comparing Sedimaging and sieving results.

The image features of indirect methods are often fed to artificial neural networks (ANN) to calculate the percentages passing or characteristic grain size (e.g. $D_{50}$). Ghalib et al. [30] used Haralick features as inputs for an ANN to predict the particle size of uniform sands. Yaghoobi et al. [26] used Fourier transforms, Gabor filters and wavelet transforms to generate inputs for a series of ANN to predict the PSD of fragmented rock material. Manashti [19] compared the results obtained with 9 feature extraction methods covering statistical, local pattern and transform-based features to predict the PSD for synthetic images of granular materials. He obtained RMSE on the percentages passing varying between 4.8 to 6.6 % when using the features as inputs for a series of ANN. Some researchers also combined direct methods of PSD determination from photographs with ANN, for example by feeding statistics of segmented areas on a photograph to an ANN for PSD determination [31].

Deep learning methods, such as a convolutional neural networks (ConvNets) [32], can be considered as indirect methods for PSD determination. ConvNets [33–37] provide a general neural network structure for inputs consisting of images, sounds, and videos. Contrarily to the classical ANN methods described previously, the ConvNet processes the image itself via a layer-wise sequence of convolutions with trained filter banks separated by non-linear activation functions (e.g. rectification), where the convolution responses ultimately serve as the inputs of a fully connected neural network. A typical ConvNet consists of millions of weight parameters (i.e. filter coefficients) across a number of layers, hence the notion of a deep ConvNet. ConvNets serve as a general computational tool in virtually any image analysis context, including medical image analysis [38–42] and speech recognition [33, 37, 43]. Well-known ConvNet architectures include AlexNet [36], SqueezeNet [44] , GoogLeNet [45], InceptionV3 [46], DenseNet201 [47], MobileNetV2 [48], ResNet18 [49], ResNet50 [50] , ResNet101 [50], Xception [51], InceptionResNetV2 [52, 53], ShuffleNet [54], and NASNetMobile [55]. These architectures are typically trained on large numbers of generic image categories and objects, e.g. the ImageNet [56] dataset consisting of 1000 different object classes, allowing them to recognize wide variety of common objects, such as computers, cups, and animal species and to generalized to new objects.

There are only a few examples of ConvNet applications for particle size analyses [32, 57–59]. Pirnia et al. [57] performed the first experiment using ConvNet for PSD determination in geotechnical engineering. Their ConvNet was also a preliminary version of PSDNet, the network that is introduced in this paper. A database containing more than 53 000 synthetic grayscale images of granular material was adopted. The images were obtained with the discrete element code YADE and had a resolution of 128×256 pixels The Microsoft Cognitive Toolkit (CNTK) [60] was used to train and test the model. The model included four convolutional layers followed by a rectified linear unit (ReLU) layer. Max pooling layers were used after the first two convolutional layers and after the final convolutional layer. Three dense layers were used after the ReLU layer. RMSE on the percentages passing of 6.9, 4.2, and 9.1 % were obtained for all sieves, only the finest sieve, and only the coarsest sieve, respectively.

Buscombe [32] introduced SediNet, a ConvNet for the size classification of sediment images. PSD obtained from a manual segmentation of the images were used as the ground truth. SediNet consists of four convolutional blocks each having several two-dimensional convolutional filter layers, batch normalization layers, and two-dimensional max pooling layers. SediNet uses RGB images of 512×512 pixels. It was trained for three regression models. In the first regression model, SediNet used 205 images to predict 9 grain size percentiles and reach an RMSE ranging from 24 to 45 % of the mean size for each percentile. The second regression model was run on 31 images of beach sand with RMSE ranging from 16 to 29 % of the mean size for each percentile. The first and second methods predicted cumulative percentage passing in pixels. The third regression model predicted the sieve size directly in mm with an RMSE (normalized by the mean grain size) of 22 %.

Mcfall et al. [59] conducted a study using four types of analyses to predict the grain size distribution of several beach sands collected by citizen scientists. They compared three image analysis techniques with sieve analyses: direct analyses based on image segmentation, transform-based features, and convolutional neural networks. The RGB images were gathered using cell-phone cameras. The final images had a size of 1024 × 1024 pixels. For the direct analysis, the grains in each image were identified by segmentation, and their size was estimated individually. Direct analysis by segmentation had a mean percentage error on $D_{50}$ of 34.2%. This percentage error was reduced to 13.0% by drying the sample and by using a black background for the photograph. The Python implementation of the wavelet feature extraction method (pyDGS) proposed by Buscombe [32] was used to estimate the grain size distribution from textural features. The mean percentage error on $D_{50}$ was 36.4%. For the ConvNet, they used the SediNet framework [32] to estimate the PSD. The mean percentage error on $D_{50}$ was 22%.

GRAINet [58] is a ConvNet model to predict the grain size distribution of gravel bars from unmanned aerial vehicles (UAV) images. Lang et al. [58] used 1,419 images from 25 gravel bars along six rivers in Switzerland with grain sizes ranging from 0.5 to 40 cm. As Buscombe [32], the ground truth consisted of PSD obtained from a manual segmentation of the images. Input images for GRAINet had a resolution of 500×200 pixels with normalized RGB channels. GRAINet includes a single 3×3 "entry" ConvNet layer, six convolutional blocks with batch normalization and RELU layers, and a 1×1 convolutional layer. Regression of the mean particle diameter ($d_m$) with this model leads to a RMSE of 27 % of the mean $d_m$ value, where $d_m = \sum P_i d_i /100$ with $d_i$ and $P_i$ as the mean size and percentage passing for bin I of the PSD.

While traditional neural networks can be trained properly with a few hundreds samples, ConvNet usually require thousands of images to be trained [61]. The main dataset used to train SediNet included only 409 images. Lang et al. [58] noted that SediNet appears to suffer from overfitting due to its small dataset. GRAINet was based on a larger dataset including 1,419 images. Different image augmentation methods can facilitate ConvNet training by increasing the number of images to avoid overfitting. Lang et al. [58] flipped their images vertically and horizontally to increase their number. Mcfall et al. [59] used 63 images and increased this number to 517 by flipping the images horizontally and by dividing each images into sub-images. Due to the resources needed to prepare large datasets for training ConvNet models, some researchers [62–64] have tried to include synthetic images in their datasets. Duhaime et al. [64] used three datasets comprising only real photographs, only synthetic images prepared with the discrete element method and a combination of both real photographs and synthetic images to train a series of ANN to predict the PSD from local entropy features. Similar features and performances were obtained with the three datasets. These results demonstrate that synthetic images can be used to generate rapidly a series of datasets simulating photographs with different characteristics (e.g. viewpoints, number of images, lighting conditions) to study their influence before creating the real dataset.

The main objective of this paper is to analyze the ability of ConvNet to predict the PSD of granular materials using synthetic images. The discrete element method was used to generate a large number of images corresponding to two viewpoints: from the top of a transparent container and from underneath. The performances of a ConvNet trained from scratch using our synthetic images was compared with the performances of generic pretrained ConvNets used as feature extractors, a process known as transfer learning. Color and grayscale images of different sizes were used to train and to evaluate the ConvNet. A large dataset of synthetic images was utilized to avoid overfitting and to show how synthetic images can be used a priori to verify the influence of different image parameters. This paper is the first to test pretrained ConvNet models for PSD determinations. Thirteen pretrained networks were compared in this study.

The dataset, the ConvNet structures, the pretrained networks, and transfer learning structure will be discussed in the methodology section. The results from each model and each dataset are compared in the result section.

## 2. Methodology

### 2.1 Preparation of synthetic granular image dataset

The discrete element code YADE was used by Pirnia et al. [65] to create synthetic images of granular materials. A virtual box was filled with spherical particles with predetermined PSD and random colors. Color images were taken from the top (Fig. 1a) and from under the transparent box (Fig. 1b). For some image analysis methods, the color images were converted to grayscale (bottom row in Fig. 1). Images were obtained for a wide range of PSD with particle sizes between 75 and 1180 µm. In terms of pixels, the particle sizes range from 12 to 106 pixels in the original 400 by 400 pixels images. The PSD were obtained by varying by 5 % increments the cumulative percentages passing for sieve sizes of 106, 150, 250, 425, and 710 µm. Each of the 53,003 top and bottom image pairs were used to obtain four views: Top (T, Fig. 1a), Under (U, Fig. 1b), Top-Under (TU, Fig. 1c), and Stretched Top-Under (STU, Fig. 1d). Images in dataset TU combine the T and U images for a size of 800×400 pixels. The STU dataset contains the TU images resized to 400×400 pixels. Dataset STU was created to feed both views at once to the pretrained models as they require square images. The horizontally scaled images (STU) introduce artificial affine deformations to the particles and textures, but these deformations should not have an influence on the results as a specific network was trained for each dataset. The T, U and STU image datasets were downscaled to 32, 64, 96, 128, and 160 pixels for our model. Images of between 224 and 331 pixels were used to feed the pretrained networks.

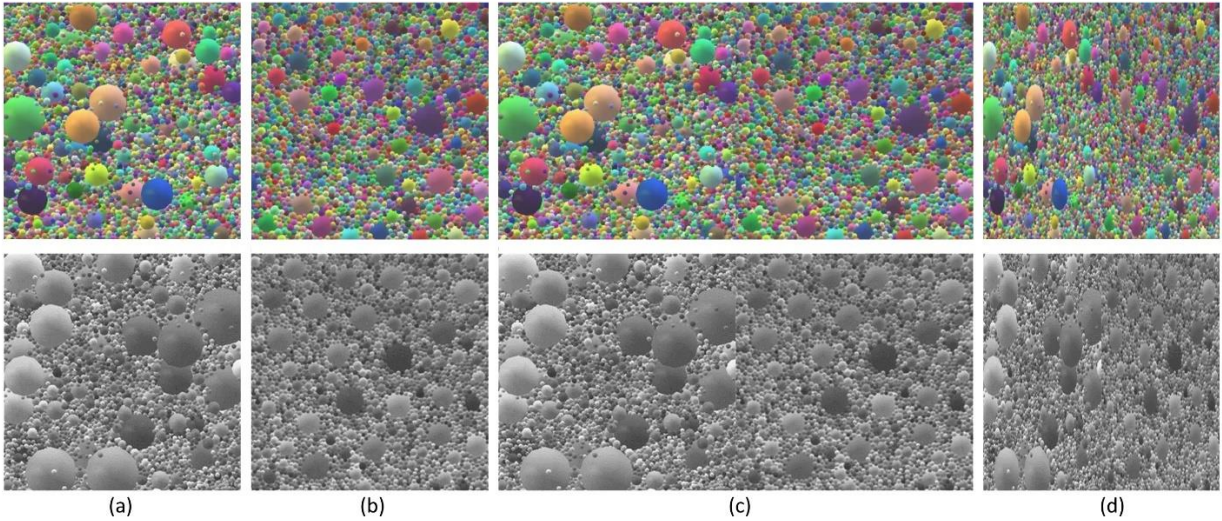

**Fig. 1** *(a) Top, (b) Under, (c) Top-Under, and (d) Stretched-Top-Under views of the virtual granular material in a transparent box. Color images are shown in the first row. Grayscale images are shown in the second row.*

## 2.2 PSDNet

ConvNet structure is centered on convolutional layers. These layers extract feature maps from the images. The main difference between the textural features of indirect methods of PSD determination and ConvNet is that the later learns the weight and bias parameters of the filter kernels during network training. Non-linear functions such as ReLU and max pooling can be applied after the convolutional layers. Finally, fully connected layers can be followed by non-linear functions, including batch normalization, ReLU, and dropout, to predict the results.

Our proposed model is called PSDNet, a regression form of ConvNet. It predicts the percentage passing for a series of sieves directly from images. Different combinations of convolutional layers with various depths and settings have been assessed to obtain the best model of PSDNet to predict PSD. Images of 32, 64, 96, 128, and 160 pixels have been examined in both grayscale and RGB color modes to evaluate the effect of color and size of input image on the accuracy of PSDNet. MATLAB 2019b was used to define the architecture of the networks and for their training in this study.

### 2.2.1 Model name

To simplify the presentation of the methodology and results, abbreviations are used for the different dataset and model names. Each model name starts with G for grayscale images or C for color images. The number that follows gives the image size (32, 64, 96, 128, 160, 224, 227, 299, and 331 pixels). The name ends with the viewpoint and image type: T (Top), U (Under), TU (Top and Under), and STU (Stretched Top and Under). For example, dataset G128T includes grayscale images taken from the top of the virtual box with a size of 128×128 pixels. For model names ending with TU, the size in the abbreviation corresponds to the image height. The image width is twice the size given in the abbreviation. For example, model C32TU was trained with color images with views from the top and from under with a resolution of 32×64 pixels. For statements that are independent of image size, model names are used without image sizes, such as CU (color images from under) or GSTU (grayscale images with the stretched top and under views).

### 2.2.2 Structure of PSDNet

PSDNet consists of four convolutional blocks each with a set of convolutional layers followed by non-linear layers, including batch normalization, ReLU, max pooling, and dropout. The convolutional blocks are followed by three blocks comprising a fully connected layer followed by batch normalization, ReLU, and dropout layers. The structure of the convolutional and fully connected blocks will be described in the following sections. The architecture of PSDNet for image dataset G128TU is shown in Fig. 2.

Convolutional layers have two main parameters: the number of filters and their size. The filter size represents the size of the kernel that sweeps the image [58]. For example, a filter size of five pixels implies that a neighborhood of 5×5 pixels is used to calculate each value in the feature map. The number of filters determines the number of independent kernels that sweep the image, each with its own set of weights and its bias. Each filter produces its own layer, or feature map. For example, 64 output layers are created by applying 64 filters on an input layer. The number of filters controls the network depth. Lower filter numbers lead to shallower networks.

SediNet and GRAINet, the previous ConvNets for PSD determination, had different architectures. Lang et al. [58] designed GRAINet based on one entry convolutional layer followed by six residual blocks (three convolutional blocks and three identity blocks). Each block contains three convolutional layers followed by batch normalization and ReLU layers. The network ends with one ConvNet layer, average pooling and SoftMax to classify the results. SediNet [32] consists of four convolutional blocks. Each block contains between 16 and 64 convolutional filters followed by batch normalization and max-pooling layers. The last convolutional block is followed by a dropout layer, a fully connected layer, and an output layer to predict the grain size percentiles. GRAINet and SediNet were respectively developed for 500×200 and 512×512 pixel color images.

PSDNet was designed based on four convolutional blocks that contain dropout layers in the second and fourth blocks. The final regression layer is preceded by three fully connected blocks before the regression output layer. Each fully connected block has a dropout ratio of 20 %. Previous models did not use dropout (GRAINet) or used it only in the fully connected block with a dropout ratio of 50 % (SediNet). Recently, dropout with a lower ratio (e.g. 20 % for PSDNet) have been used in the convolutional layer to avoid overfitting and increase model accuracy [66]. The PSDNet structure was optimized for multi dimensional input images from 32 up to 160 pixels for both color and grayscale images.

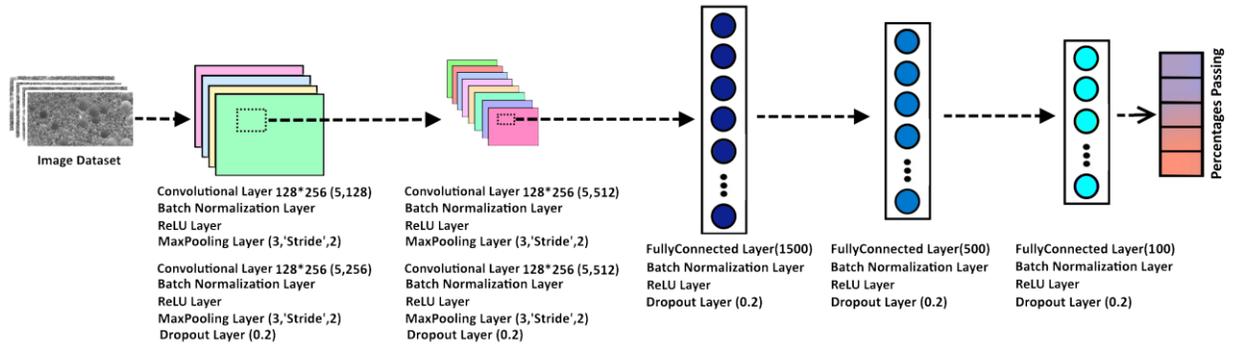

**Fig. 2** PSDNet structure for G128TU. Four blocks of convolution layers are followed by batch normalization, ReLU, max poolling, and dropout layers. The convolutional blocks are followed by three fully connected layers and by batch normalization and ReLU layers.

The influence of the number of filters on the performances of PSDNet was verified by training a series of models with different numbers of filters in the four convolutional blocks. Models G32STU, C32STU, and G32TU were chosen for the comparison. The number of filters in the first convolutional block was varied between 4 and 256. As shown in Fig. 3, more than 32 filters were needed in the first block to reach the plateau corresponding to the lowest RMSE values on the percentages passing for models G32STU and C32STU. Similarly, a filter number equal to the input image size gave the best results for models with larger images (not shown here). Consequently, the filter number of the first layer was set equal to the input size. Larger numbers of filters could be used, but it increases the size of PSDNet and memory requirements for the Graphics Processing Unit (GPU), especially for larger images.

Preliminary tests with PSDNet have shown that better performances were obtained when the filter number was increased deeper in each network. Better performances were obtained when the number of filters was set as the image size for the first convolutional layer, double the image size for the second convolutional layer, and quadruple the image size for the third and fourth layers. For example, the number of filters for an image input size of 128 pixels was set at 128, 256, 512, and 512 for convolution blocks 1 through 4 (Fig. 2). Buscombe [32] also used four blocks with increasing filter numbers for SediNet (16, 32, 48, and 64 for the first to last blocks).

Fig. 4 shows the effect of filter size on the performances of three models. The filter sizes with the minimum RMSE on the percentages passing are 3, 4, and 5. Filters of 5×5 pixels were selected for the PSDNet convolutional layers. A padding of size 2 was applied around the inputs (image and feature maps) to maintain a constant output size. The inputs were padded with zeros.

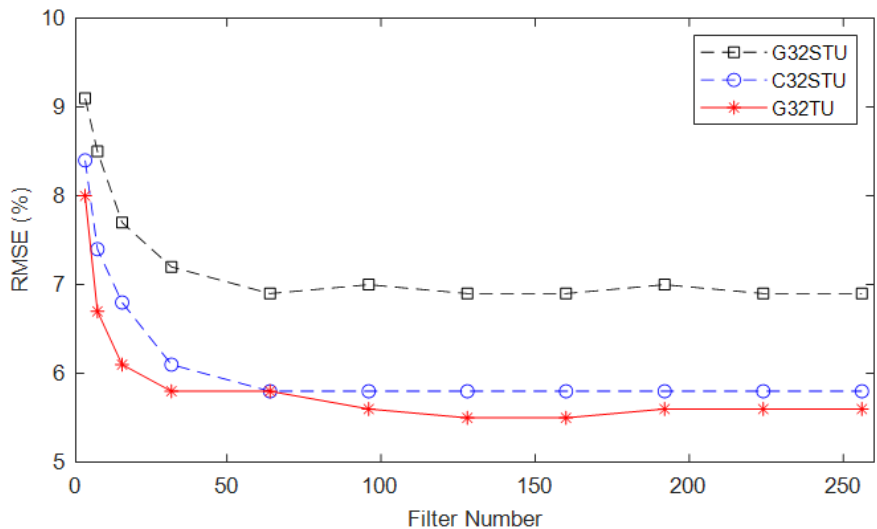

**Fig. 3** *Effect of filter number on the percentage passing RMSE when filter size is set to five.*

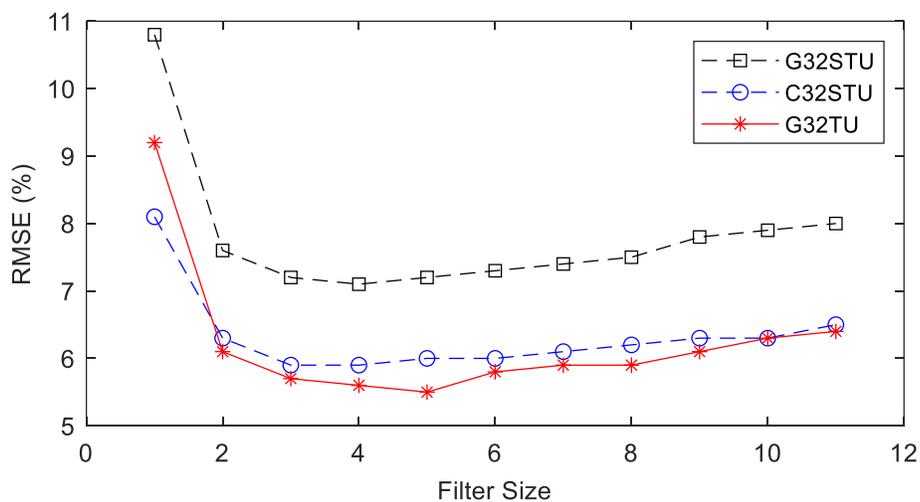

**Fig. 4** *Effect of filter size on percentage passing RMSE when filter number is set to 96, 32, and 32 for models G32TU, G32STU, and C32STU, respectively.*

Batch normalization layers set the mean and variance of the output of the previous layer [61]. This function can be used to speed up training and reduce the sensitivity to network initialization data [52]. Typically, batch normalization is used after each convolutional layer and before other non-linear functions such as the Rectified Linear Unit (ReLU) layer.

ReLU layers are used regularly after batch normalization layers [67]. These layers intensify the model non-linearity by replacing all the negative activations with zero [61]. The activation function of ReLU layer is defined as:

$$f(x) = \max(0, x) \quad (1)$$

where f($x$) is the threshold function, and $x$ is the function input.

Max pooling layers [68] are added to the network to reduce the data size, computation, and overfitting [32]. Max pooling separates the input layer into smaller pooling areas and finds the maximum activation for each area. Max pooling is applied to $n \times n$ pooling areas. In PSDNet, the max pooling layer is applied with a 3×3 filter size and a 2×2 stride. Stride is the movement of the pooling region in both vertical and horizontal directions. A stride of more than 1 reduces the size of the layer output with respect to the input. In PSDNet, the pooling regions overlap because the stride is less than the pooling dimension.

Dropout layers are used after every two packages of convolutional block, including convolutional layer, batch normalization, ReLU, and max pooling. To avoid overfitting of training data, dropout layers can be used to change some of the activations to zero randomly [66, 69]. In PSDNet, dropout layers were used so that 20 % of the number of input layers was set to zero.

Next, three fully connected (dense) layers followed by ReLU and dropout layers are applied. Fully connected layers function like normal neural networks and connect all activations from the previous layer [70]. The last layer in PSDNet is linear. This layer provides a regression output to predict the percentages passing for each sieve.

The total number of learnable parameters for PSDNet varied from 1.6 million for G32T up to 182 million for C160TU. As a comparison, GRAINet [58], ShuffleNet [54], ResNet101 [50] and vgg19 [71] including 1.6 million, 14 million, 44 million, and 144 million parameters, respectively.

### 2.2.3 Network training

For training PSDNet, Stochastic Gradient Descent with Momentum (SGfDM) was used as a solver. The best PSDNet results were obtained when the initial learning rate was set at 0.0001. Different minibatch sizes were used based on the image size and GPU memory. For example, the minibatch size was set at 256 for G32T and 10 for C160TU. The minibatch size was limited by GPU memory (NVIDIA GeForce RTX 2080 with 8 GB memory and NVIDIA GeForce GTX 1650 with 4 GB memory). The minibatch size corresponds to the number of images that are used to calculate the gradient of the loss function and to update the weight. The training and validation datasets were shuffled before each training epoch. Due to the large training (42 403 images) and validation (5300 images) datasets, the training and validation process took about 1 hour for grayscale images with a size of 32 pixels and up to 100 hours for the color images of 160 pixels. The complete dataset was divided in training, validation and testing subdatasets according to proportions of 80/10/10 %, respectively. Percentages passing were predicted for five sieves (106, 150, 250, 425, and 710 μm).

Figure 5 shows the effect of the number of training epochs on the RMSE of the test dataset. For C128T, the best RMSE on the percentages passing is obtained when the epoch is equal to 7.

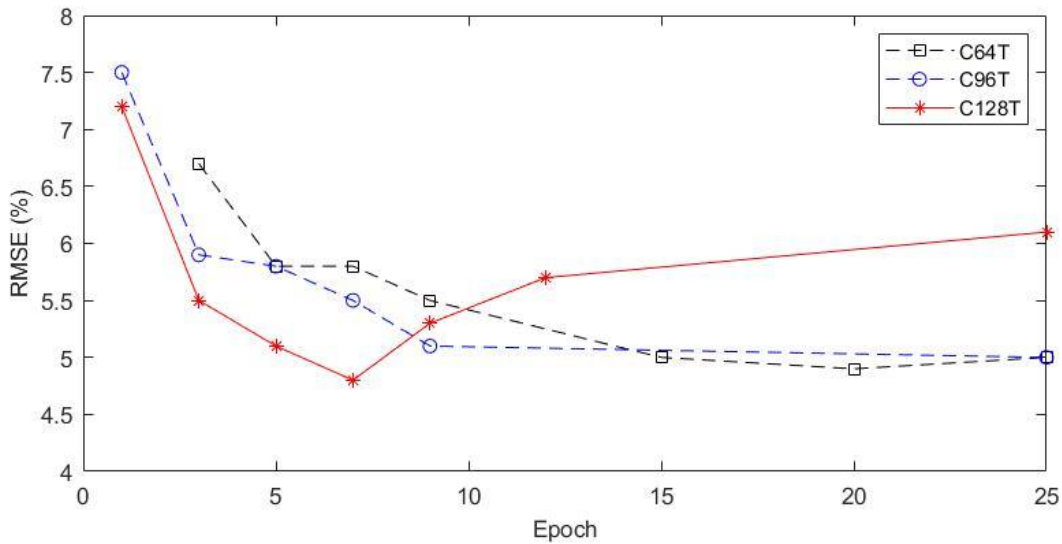

**Fig. 5** *Effect of number of epochs on the RMSE of the percentage passing for all sieves and the test dataset*

## 2.3 Pretrained ConvNets

Organizing thousands of labeled images is not possible for all applications as such datasets can be rare, expensive, or inaccessible [72]. In this case, high-performance models can be pretrained with an accessible dataset for a different application and used for transfer learning. For example, networks trained on the ImageNet [56] dataset for the ImageNet Large-Scale Visual Recognition Challenge (ILSVRC) [73] can be partially retrained or used as feature extractors for other applications and datasets.

### 2.3.1 Feature extraction

The feature extraction method is a quick way to use pretrained ConvNet. It does not require extensive computational power. This method is more useful when the dataset size is small. In the feature extraction method, the pretrained networks are used to extract features from the last layer. These features are fed to a new ANN to predict the percentages passing. With this method, the original pretrained network is not trained with the granular material images.

Twelve networks were compared to evaluate the performances of pretrained networks as feature extractors to estimate the PSD of granular materials. The networks include AlexNet, SqueezeNet, GoogLeNet, InceptionV3, DenseNet201, MobileNetV2, ResNet18, ResNet50, ResNet101, InceptionResNetV2, ShuffleNet, and NASNetMobile. The last layer of each pretrained network was used to extract 1000 features for the T, U, and STU images, separately. For the TU images, 2000 features were extracted by combining the features from the T and U images. The size of the input images for these networks ranges between 224 and 331 pixels depending on their structure. It should be noted that the input size of the pretrained networks is larger than the input size used with PSDNet (32 to 160 pixels).

The PSD was predicted with a neural network with one hidden layer with 10 neurons. The ANN was trained with MATLAB. For the training function, the Levenberg-Marquardt algorithm was set. The complete dataset was divided between the training, validation and testing subdatasets with proportions of 70/15/15 %, respectively.

### 2.3.2 Transfer learning

With transfer learning, the last layer of the pretrained networks can be replaced by a new regression layer and the models can be trained again with the new dataset [72]. This method is more computationally intensive than feature extraction as the complete pretrained networks are trained starting from the weight obtained during training with the primary dataset (e.g. ImageNet). Nevertheless, transfer learning is a faster than designing and training a new ConvNet model for large datasets. In this study, 12 pretrained networks including AlexNet, GoogLeNet, InceptionV3, DenseNet201, MobileNetV2, ResNet18, ResNet50, ResNet101, Xception, InceptionResNetV2, ShuffleNet, and NASNetMobile were trained to predict the PSD.

The classifier layer (softmax) was replaced with a regression layer to predict the PSD. The networks were trained for 5 epochs. Images from the T, U and STU datasets were used to train the networks. Since the ConvNet were pretrained for square images, it was not possible to evaluate the TU dataset with this method.

## 3. Results and discussion

### 3.1 Influence of image size and color

Images of 32, 64, 96, 128, and 160 pixels were examined in both grayscale (G) (Fig. 6, top row) and RGB (C) (Fig. 6, bottom row) modes to understand the effect of image input size and color on the ConvNet results. Four viewpoints and image stitching methods were compared: top (T), under (U), top and under (TU), and stretched top and under (STU).

As shown in Fig. 6, the RMSE on the percentages passing for all sieves generally decreases when the image size increases from 32 to 160 pixels in both grayscale and color modes. The image combining the top and under viewpoints has the best results for both color and grayscale images, followed by stretched top and under, top, and under views, respectively. For example, the RMSE for the GTU datasets goes from 5.8 to 3.1 % for 32 and 160 pixels images, while for GU it goes from 8.4 to 4.8 % for the same sizes. It is worth noting that the size of TU images is doubled compared to other methods. STU images achieves better results compared to the top or under views with the same size, even if the images are stretched.

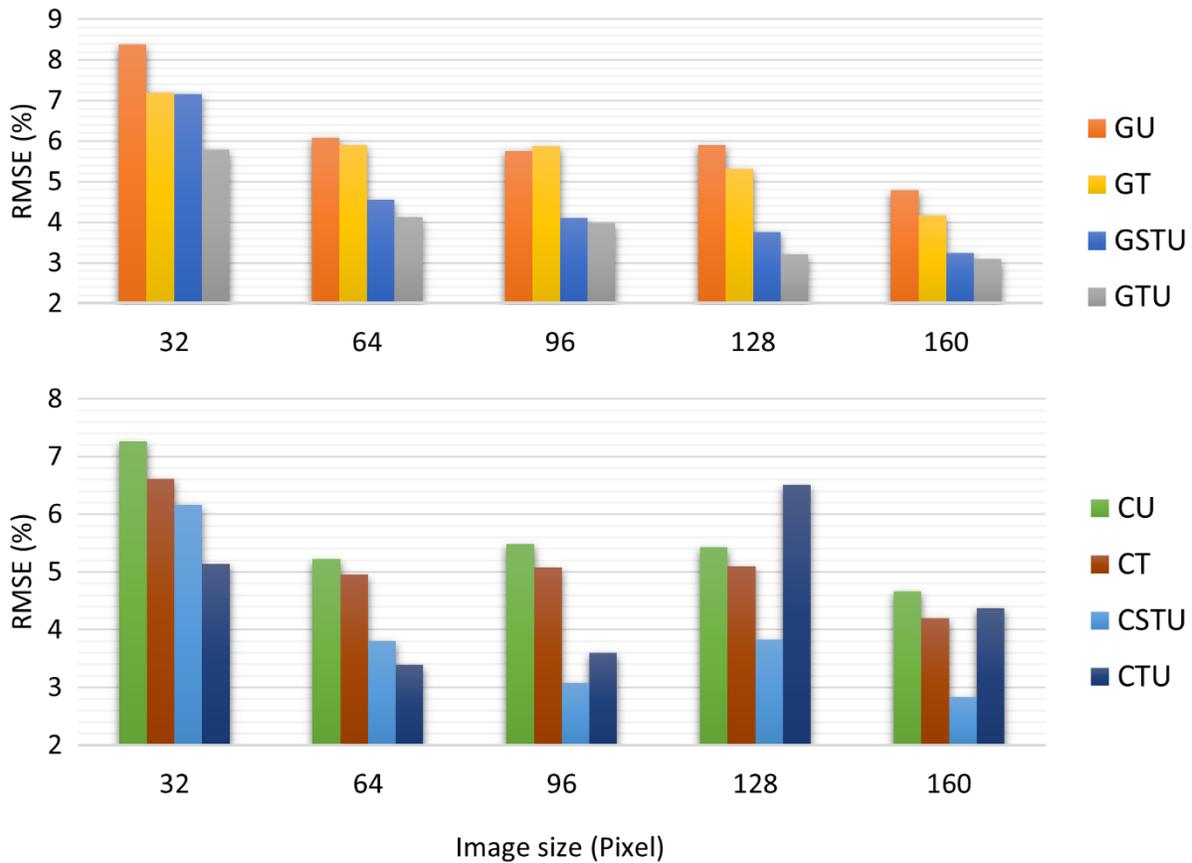

**Fig. 6** *Effect of PSDNet image size on RMSE in grayscale (top row) and color (bottom row) images for different capture views of Under (U), Top (T), Stretched Top-Under (STU), and Top-Under (TU) datasets*

Most of the time, the RMSE is lower when using color images compared to grayscale images. The RMSE reaches 2.8% for C160STU. Color datasets can reach better results than grayscale datasets with smaller images. For example, model C96STU has a similar RMSE to model G160TU. As for the grayscale images, TU models have better results followed by STU, T, and U for RGB views, respectively. The RMSE difference between grayscale and color modes is reduced by increasing the image size. For example, C32U is better than G32U by more than 1%, but C160U is only 0.1% better than G160U.

The previous ConvNet model trained by Pirnia et al. [57] achieved RMSE on the percentages passing of 6.9, 4.2, and 9.1 % for all sieves, the finest sieve, and the coarsest sieve, respectively. As mentioned previously, it was trained using grayscale images of 128×256 pixels. With the same dataset, PSDNet could reach RMSE values of 3.2, 1.8, and 3.0 % for all sieves, the finest sieve, and the coarsest sieve, respectively. The version of PSDNet trained with datasets C128STU and C160STU had an RMSE value of 1.6% for fine particles and 2.8 % for all sieves, respectively. This highlights the importance of optimizing the network architecture and parameters.

The influence of using a synthetic dataset for ConvNet training instead of real photographs is difficult to appraise. Duhaime et al. [64] compared the RMSE on the percentage passing for a series of ANN trained using local entropy features and different datasets with the same PSD in terms of pixels per diameter. They obtained a RMSE of 3.4 % for synthetic images similar to the grayscale images used in this paper, 3.8 % for real photographs and 4.4 % when combining both datasets. Therefore, even if better performances can be expected from synthetic dataset, the difference with real photographs is relatively small. It should be noted that synthetic datasets have been used for ConvNet training in other fields (e.g. [62, 74]).

Manashti [19] used the same synthetic images to compare nine indirect methods based on traditional features used as inputs for a series of ANN. The best method achieved RMSE values of 3.4, 1.7, and 4.7 % for all sieves, the finest sieve, and the coarsest sieve, respectively. In comparison, PSDNet produced RMSE values of 3.2, 1.8, and 3.0 % for the same dataset (G128TU). Thus, methods based on traditional feature extraction obtained similar results for finer particles, but better results were obtained with ConvNet for larger particles. Most methods based on traditional feature extraction performed relatively poorly for coarser particles. This could be a significant advantage of ConvNet.

PSDNet achieved good results compared to those reported in the literature for real datasets and direct methods based on image segmentation. RMSE of respectively 13-36 % and 15-20 % can be calculated from the results presented by Liu and Tran [10] and Sudhakar et al. [75], respectively. Both studies were based on the commercial software package WipFrag. It should be noted that training and running ConvNet requires much more computational power and much larger datasets than the conventional direct and indirect methods. Training ConvNet with a thousand images or less is likely to lead to overfitting problems [58]. With smaller number of images, the direct method used in commercial software, indirect methods based on feature extractions, and pretrained ConvNet are more suitable. Combining synthetic and real datasets as done by Duhaime et al. [64] might be an efficient solution to train ConvNet networks with only a few thousands real photographs.

## 3.2 Pretrained feature extraction

Figure 7 presents the results for pretrained ConvNet used as feature extractors with the TU dataset. InceptionResNetV2 achieved the best results with an RMSE value of 3.6 % for all sieves. In addition, a combination of the features extracted from the MobileNetV2, ResNet101, DenseNet201, and InceptionResNetV2 networks (8000 features) reached a RMSE value of less than 3 % for fine particles.

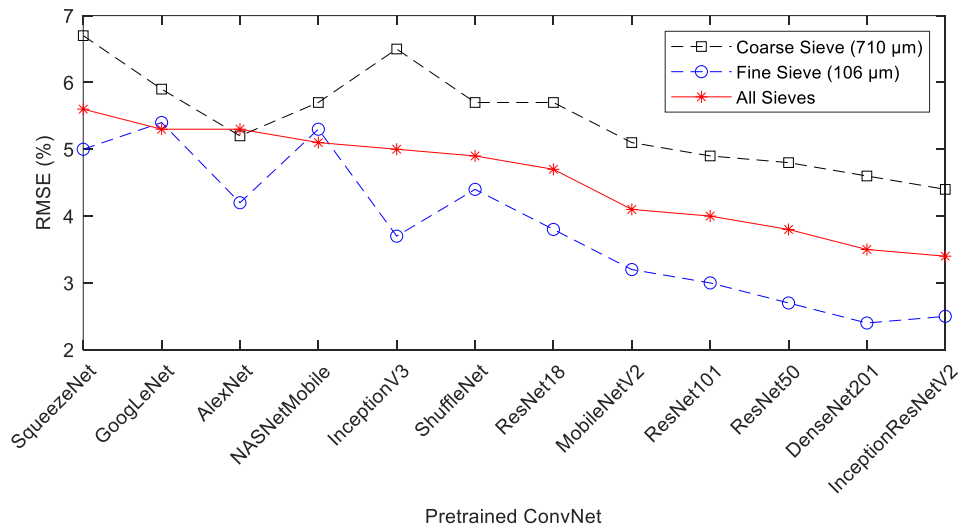

**Fig. 7.** *Performances of pretrained ConvNet models used as feature extractors on the prediction of all sieves, coarse, fine sieves for TU view*

Fig. *8* represents the performances of different pretrained models used as features extractors to predict the PSD for the T, U, TU, and STU datasets. Like PSDNet, pretrained ConvNet also achieved better results for TU images followed by STU, T, and U. InceptionResNetV2 achieved an RMSE value of 3.4 % for TU images as the best result. The highest RMSE was obtained with SqueezeNet (7.5 %) for the U dataset. The pretrained models used the color images of 224 to 331 pixels. Compared to PSDNet, the best pretrained feature extractor achieved results similar to network C64TU. Network G128TU achieved better results than InceptionResNetV2 for TU images.

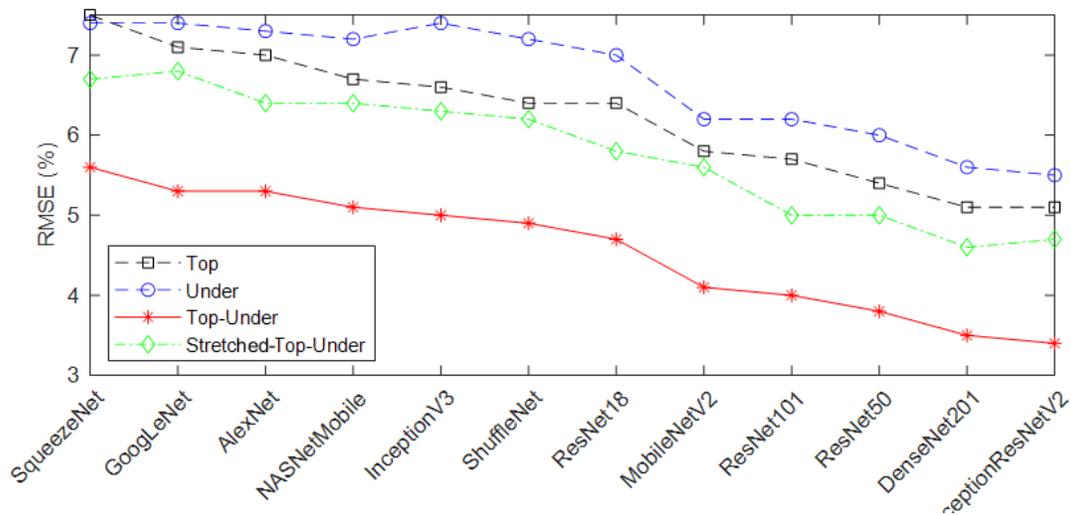

**Fig. 8.** *Effect of image viewpoint and stitching on PSD prediction with 12 pretrained models used as feature extractors*

The results for pretrained feature extractors are similar to the results reported by Manashti [19] for nine different traditional feature extractors. For fine particles, DenseNet201 predicts the percentages passing with an RMSE of 2.4 % while LCP got an RMSE of 1.9% as the best traditional feature extractor. The pretrained models predict the PSD for the TU dataset with an RMSE value ranging from 3.4 to 5.6 %, while the RMSE values for traditional feature extraction ranges from 3.4 % to 6.9 % when all sieves are considered. It should be noted that the pretrained models reach these results using 1000 or 2000 features whereas the traditional models use between 7 and 3880 features. In comparison, Liu and Tran [10], achieved RMSE values ranging from 13 to 36% for backfill muck pile with FragScan, WipFrag, and Split. Sudhakar et al. [75], in other projects, reached an RMSE value of 15 % for WipFrag and 20 % for Fragalyst for the blasted sandstones. Both papers by Liu and Tran [10] and Sudhakar et al. [77] used real photographs.

### 3.3 Transfer learning

With the transfer learning method, the last classification layer of the pretrained ConvNet models was replaced with a regression layer. The model parameters were fine-tuned starting from the weight and bias values learnt during pretraining with the ImageNet database. Figure 9 shows the RMSE on the percentages passing for all sieves after training for five epochs with the T, U and STU datasets. Transfer learning models achieved their best results with the STU views with RMSE values ranging from 3.6 to 5.7 %. In contrast with the feature extraction method presented in the previous section, similar results were obtained with the T and U datasets, with the U dataset performing better for half of the pretrained models.

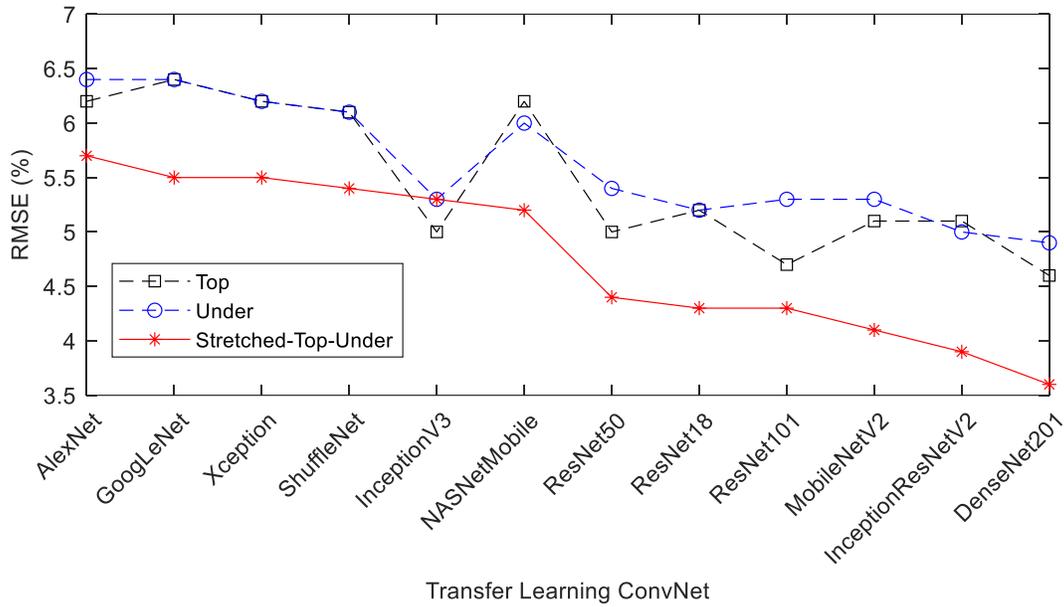

**Fig. 9** *Results of transfer learning of 14 pretrained ConvNet models for T, U, and STU datasets*

Compared to the feature extraction method, the transfer learning method gives better results when comparing the results for the same dataset. On the other hand, transfer learning cannot run on the TU dataset, while feature extraction achieves good results on the same dataset. InceptionReseNetV2 as a feature extraction method achieved an RMSE of 3.4 % for the Top-Under view, while DenseNet201, the best method of transfer learning, achieved an RMSE of 3.6 % for the STU view.

The results presented by Manashti [19] for a selection of 618 traditional features are slightly better than the results obtained with transfer learning in this paper. Some of the transfer learning methods presented in this paper have better results than the individual feature types compared by Manashti [19] (HOG, Wavelet, or Fourier) with RMSE values ranging from 6.2 to 6.9 % for all sieves.

### 3.5 PSD chart

Fig. *10* compares the real and predicted percentages passing for a sample of six randomly selected synthetic soil images from the C160STU dataset. The percentages passing are plotted on the usual semi-log plot. The blue diamonds and lines indicate real PSD, while the red circles and lined indicate the predictions. Overall, the real and predicted grain size distributions are similar. It should be noted that PSDNet sometimes predicts PSD that are physically impossible. For example, in Fig. 10e, the percentage passing decreases from 425 to 710 μm.

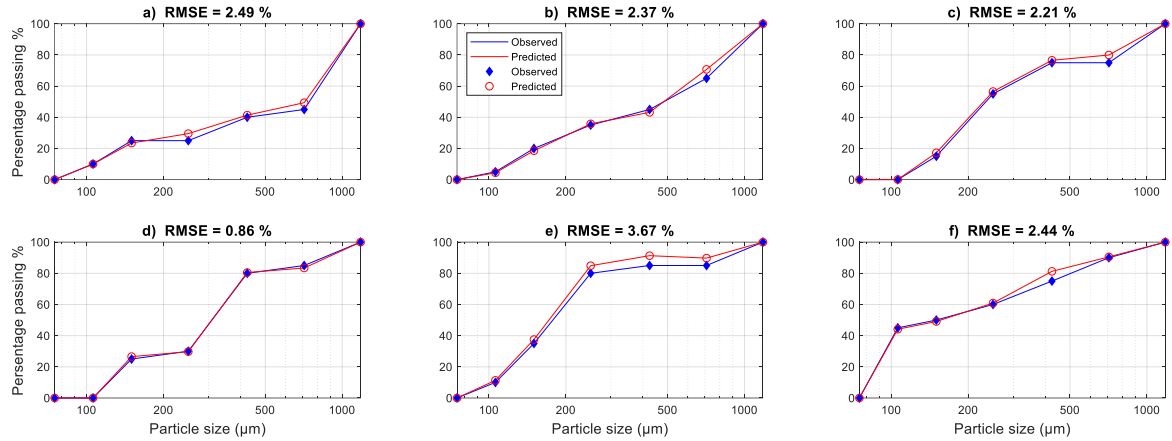

**Fig. 10** *Six randomly selected samples of real and predicted PSD obtained with PSDNet for dataset C160STU.*

## 3.6 $D_{50}$

The $D_{50}$ can be estimated using a linear interpolation between the two PSD data points below and above a percentage passing of 50%. Figure 11 represents the prediction and observation of the percentage error of $D_{50}$ for the C96STU dataset. The mean error percentages on $D_{50}$ for model C96STU was 5.6%. Slightly higher error percentages were obtained for larger images with transfer learning (6.0%) and pretrained feature extraction (6.0%). Quite similar results on error percentages were achieved for traditional feature extraction (6.1%)[19]. Mcfall et al. [59] obtained an error of 22 % on the $D_{50}$ of beach sand with SediNet, the ConvNet presented by Buscombe [32]. For rock fragmentation, the mean $D_{50}$ error was 55 % and 100 % for WipFrag and Split, and FragScan, respectively [10].

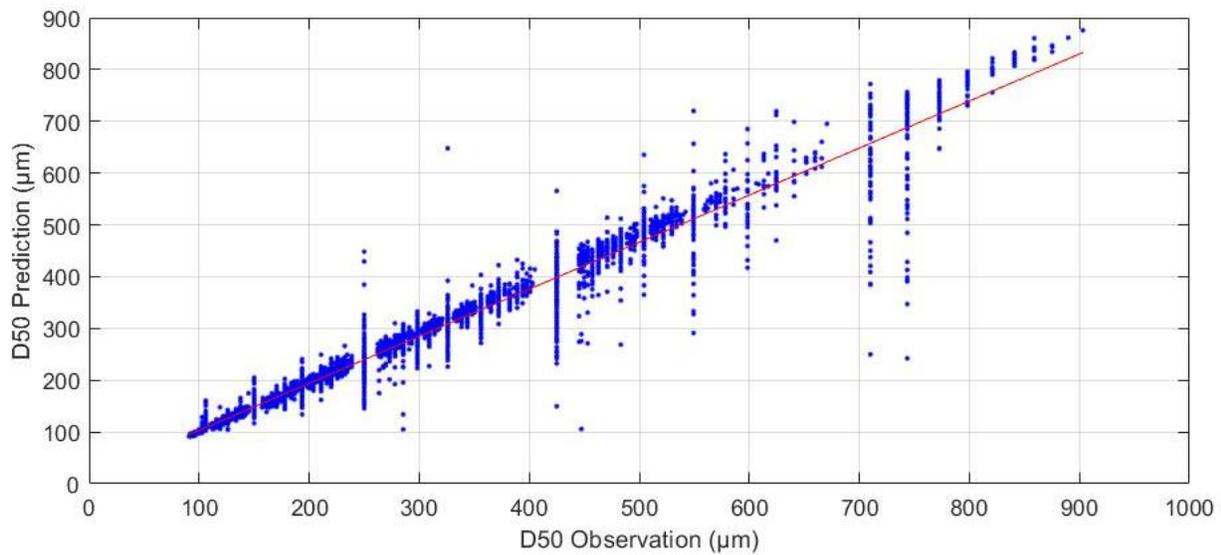

**Fig. 11** *Predicted and observed $D_{50}$ for model C96STU*

## 4. Conclusion

The primary aim of this study was to use ConvNet to predict the PSD of granular material using a large dataset of synthetic images. The results show the ability of ConvNet to predict the PSD from images even though some particles are fully or partly covered by others. This new method can provide fast, accurate, and economical online PSD predictions in the laboratory or in the field.

The best performances were obtained with PSDNet (model C160STU) with RSME values of 1.8 % for fine particles, 3.3 % for coarse particles, and 2.8 % for all sieves. Manashti [19] presented results for the same dataset with traditional feature extraction methods. His combined feature extraction method reached RMSE values of 1.7, 4.7, and 3.4 % for fine particles, coarse particles and all sieves, respectively. The performance improvement with ConvNet for coarser sieves is significant as traditional feature extraction tended to perform poorly for coarse particles.

This paper was the first to use pretrained ConvNet for feature extraction or transfer learning to determine the PSD of granular material. Several pretrained models were examined in this paper using our synthetic image dataset. For feature extraction of pretrained ConvNet, an RMSE value of 3.4 % was obtained for all sieves and the STU dataset. This RMSE is identical to the value obtained by Manashti [19] for the same dataset and a combination of traditional feature extraction methods. Transfer learning did not improve the results compared to pretrained networks used as feature extractors or traditional feature extraction [19].

The role played by training is one of the main differences between direct methods of PSD determination based on segmentation and indirect methods based on textures and neural networks. Training allows the network to take into account the hidden particles implicitly. Methods based on segmentation only analyze the particles at the surface and must account for hidden particles explicitly through statistical relationships. Because they are machine learning models, ConvNet can also be improved through operation by expanding the dataset.

## 5. Acknowledgments

Training of the preliminary version of PSDNet introduced by Pirnia et al. [57] was made possible by a Microsoft Azure sponsorship. This project was supported by Hydro-Québec and the National Sciences and Engineering Research Council of Canada (NSERC).

# 6. References

1. Gee GW, Bauder JW (1986) Particle-size analysis. In: Methods of soil analysis: Part 1—Physical. Soil Science Society of America, American Society of Agronomy, pp 383–411

2. Ko Y-D, Shang H (2011) A neural network-based soft sensor for particle size distribution using image analysis. Powder Technol 212:359–366. https://doi.org/10.1016/j.powtec.2011.06.013

3. Schneider CL, Neumann R, Souza AS (2007) Determination of the distribution of size of irregularly shaped particles from laser diffractometer measurements. Int J Miner Process 82:30–40. https://doi.org/10.1016/j.minpro.2006.09.011

4. Costodes VCT, Mausse CF, Molala K, Lewis AE (2006) A simple approach for determining particle size enlargement mechanisms in nickel reduction. Int J Miner Process 78:93–100. https://doi.org/10.1016/j.minpro.2005.09.001

5. Ko Y-D, Shang H (2011) Time delay neural network modeling for particle size in SAG mills. Powder Technol 205:250–262. https://doi.org/10.1016/j.powtec.2010.09.023

6. Ilonen J, Juránek R, Eerola T, Lensu L, Dubská M, Zemčík P, Kälviäinen H (2018) Comparison of bubble detectors and size distribution estimators. Pattern Recognit Lett 101:60–66. https://doi.org/10.1016/j.patrec.2017.11.014

7. Caputo F, Clogston J, Calzolai L, Rösslein M, Prina-Mello A (2019) Measuring particle size distribution of nanoparticle enabled medicinal products, the joint view of EUNCL and NCI-NCL. A step by step approach combining orthogonal measurements with increasing complexity. J Control Release 299:31–43. https://doi.org/10.1016/j.jconrel.2019.02.030

8. Allen T (2003) Powder sampling and particle size determination. Elsevier, Amsterdam

9. Thurley MJ, Ng KC (2008) Identification and sizing of the entirely visible rocks from a 3D surface data segmentation of laboratory rock piles. Comput Vis Image Underst 111:170–178. https://doi.org/10.1016/j.cviu.2007.09.009

10. Liu Q, Tran H (1996) Comparing systems - Validation of FragScan, WipFrag and Split. In: Proceedings of the Fifth Inter- national Symposium on Rock Fragmentation by Blasting – FRAGBLAST. pp 151–155

11. Maerz NH, T.C. Palangio, T.W. Palangio, K. Elsey (2007) Optical sizing analysis of blasted rock: lessons learned. In: Materiały konferencyjne 4th EFEE World Conference of Explosives and Blasting in Vienna. pp 75–83

12. Maerz NH, Palangio TC, Franklin JA (1996) WipFrag image based granulometry system. In: Proceedings of the Fifth International Symposium on Rock Fragmentation by Blasting – FRAGBLAST. pp 91–99

13. Girdner KK, Kemeny JM, Srikant A, McGill R (1996) The split system for analyzing the size distribution of fragmented rock. In: Proceedings , FRAGBLAST-5 Workshop on Measurement of Blast Fragmentation, edited by J. A. Franklin and T. Katsabanis, 101-108. Rotterdam, NL: Balkema. pp 101–108

14. Detert M, Weitbrecht V (2012) Automatic object detection to analyze the geometry of gravel grains--a free stand-alone tool. In: River flow. Taylor & Francis Group London, pp 595–600

15. Tian DP (2013) A review on image feature extraction and representation techniques. Int J Multimed Ubiquitous Eng 8:385–395

16. Haralick RM, Shanmugam K, Dinstein I (1973) Textural Features for Image Classification. IEEE Trans Syst Man Cybern SMC-3:610–621. https://doi.org/10.1109/TSMC.1973.4309314

17. Khellaf A, Dupoisot H, Beghdadi A (1991) Entropic Contrast Enhancement. IEEE Trans Med Imaging 10:589–592. https://doi.org/10.1109/42.108593

18. Gonzalez RC, Woods RE (2002) Digital Image Processing, 2nd Edition. Pearson Education India

19. Manashti MJ (2021) Grain Size Analyses of Soils Based on Image Analysis Techniques and Machine Learning. Dissertation, École de Technologie Supérieure

20. Dalal N, Triggs B (2005) Histograms of Oriented Gradients for Human Detection. In: 2005 IEEE Computer Society Conference on Computer Vision and Pattern Recognition (CVPR'05). IEEE, pp 886–893

21. Ojala T, Pietikainen M, Maenpaa T (2002) Multiresolution gray-scale and rotation invariant texture classification with local binary patterns. IEEE Trans Pattern Anal Mach Intell 24:971–987. https://doi.org/10.1109/TPAMI.2002.1017623

22. Guo Y, Zhao G, Pietikäinen M (2011) Texture Classification using a Linear Configuration Model based Descriptor. In: Procedings of the British Machine Vision Conference 2011. British Machine Vision Association, pp 119.1-119.10

23. Zhenhua Guo, Lei Zhang, Zhang D (2010) A Completed Modeling of Local Binary Pattern Operator for Texture Classification. IEEE Trans Image Process 19:1657–1663. https://doi.org/10.1109/TIP.2010.2044957

24. Shin S, Hryciw RD (2004) Wavelet Analysis of Soil Mass Images for Particle Size Determination. J Comput Civ Eng 18:19–27. https://doi.org/10.1061/(ASCE)0887-3801(2004)18:1(19)

25. Hryciw RD, Ohm H-S, Zhou J (2015) Theoretical Basis for Optical Granulometry by Wavelet Transformation. J Comput Civ Eng 29:04014050. https://doi.org/10.1061/(ASCE)CP.1943-5487.0000345

26. Yaghoobi H, Mansouri H, Ebrahimi Farsangi MA, Nezamabadi-Pour H (2019) Determining the fragmented rock size distribution using textural feature extraction of images. Powder Technol 342:630–641. https://doi.org/10.1016/j.powtec.2018.10.006

27. Szeliski R (2011) Computer Vision, Algorithms and Applications. Springer, London

28. Tuceryan M, Jain AK (1993) Texture Analysis. In: Handbook of Pattern Recognition and Computer Vision. WORLD SCIENTIFIC, pp 235–276

29. Ohm H-S, Hryciw RD (2014) Size Distribution of Coarse-Grained Soil by Sedimaging. J Geotech Geoenvironmental Eng 140:04013053. https://doi.org/10.1061/(ASCE)GT.1943-5606.0001075

30. Ghalib AM, Hryciw RD, Shin SC (1998) Image Texture Analysis and Neural Networks for Characterization of Uniform Soils. Comput Civ Eng 671–682

31. Hamzeloo E, Massinaei M, Mehrshad N (2014) Estimation of particle size distribution on an industrial conveyor belt using image analysis and neural networks. Powder Technol 261:185–190. https://doi.org/10.1016/j.powtec.2014.04.038

32. Buscombe D (2020) SediNet: a configurable deep learning model for mixed qualitative and quantitative optical granulometry. Earth Surf Process Landf 45:638–651. https://doi.org/10.1002/esp.4760

33. LeCun, Y, Bengio Y (1995) Convolutional networks for images, speech, and time series. Handb brain theory neural networks 3361(10)

34. Chellapilla K, Puri S, Simard P (2006) High Performance Convolutional Neural Networks for Document Processing. In: Tenth International Workshop on Frontiers in Handwriting Recognition. Suvisoft, La Baule (France), pp 1–6

35. Kavukcuoglu K, Sermanet P, Boureau Y, Gregor K, Mathieu M, Cun YL, LeCun Y (2010) Learning Convolutional Feature Hierarchies for Visual Recognition. Adv Neural Inf Process Syst 23 1090–1098

36. Krizhevsky A, Sutskever I, Hinton GE (2012) ImageNet classification with deep convolutional neural networks. In: Advances in neural information processing systems. pp 1097–1105

37. Abdel-hamid O, Jiang H, Penn G, Mohamed A, Jiang H, Penn G (2012) Applying Convolutional Neural Networks concepts to hybrid NN-HMM model for speech recognition. In: 2012 IEEE International Conference on Acoustics, Speech and Signal Processing (ICASSP). pp 4277–4280

38. Lee J-G, Jun S, Cho Y, Lee H, Kim GB, Seo JB, Kim N (2017) Deep Learning in Medical Imaging: General Overview. Korean J Radiol 18:570–584. https://doi.org/10.3348/kjr.2017.18.4.570

39. Tajbakhsh N, Shin JY, Gurudu SR, Hurst RT, Kendall CB, Gotway MB, Liang J (2016) Convolutional Neural Networks for Medical Image Analysis: Full Training or Fine Tuning? IEEE Trans Med Imaging 35:1299–1312. https://doi.org/10.1109/TMI.2016.2535302

40. Yan K, Lu L, Summers RM (2018) Unsupervised body part regression via spatially self-ordering convolutional neural networks. In: 2018 IEEE 15th International Symposium on Biomedical Imaging (ISBI

2018). IEEE, pp 1022–1025

41. Hall SA, Bornert M, Desrues J, Pannier Y, Lenoir N, Viggiani G, Bésuelle P (2010) Discrete and continuum analysis of localised deformation in sand using X-ray μCT and volumetric digital image correlation. Géotechnique 60:315–322. https://doi.org/10.1680/geot.2010.60.5.315

42. Plautz T, Boudreau R, Chen J-H, Ekman A, LeGros M, McDermott G, Larabell C (2017) Progress Toward Automatic Segmentation of Soft X-ray Tomograms Using Convolutional Neural Networks. Microsc Microanal 23:984–985. https://doi.org/10.1017/S143192761700558X

43. Abdel-hamid O, Deng L, Yu D (2013) Exploring Convolutional Neural Network Structures and Optimization Techniques for Speech Recognition. In: 14th Annual Conference of the International Speech Communication Association, (Interspeech 2013). pp 73–5

44. Iandola FN, Han S, Moskewicz MW, Ashraf K, Dally WJ, Keutzer K (2016) SqueezeNet: AlexNet-level accuracy with 50x fewer parameters and <0.5MB model size. arXiv Prepr arXiv160207360

45. Szegedy C, Wei Liu, Yangqing Jia, Sermanet P, Reed S, Anguelov D, Erhan D, Vanhoucke V, Rabinovich A (2015) Going deeper with convolutions. In: 2015 IEEE Conference on Computer Vision and Pattern Recognition (CVPR). IEEE, pp 1–9

46. Szegedy C, Vanhoucke V, Ioffe S, Shlens J, Wojna Z (2016) Rethinking the Inception Architecture for Computer Vision. In: IEEE Conference on Computer Vision and Pattern Recognition (CVPR). IEEE, pp 2818–2826

47. Huang G, Liu Z, Van Der Maaten L, Weinberger KQ (2017) Densely Connected Convolutional Networks. In: IEEE Conference on Computer Vision and Pattern Recognition (CVPR). IEEE, pp 2261–2269

48. Sandler M, Howard A, Zhu M, Zhmoginov A, Chen L-C (2018) MobileNetV2: Inverted Residuals and Linear Bottlenecks. In: IEEE Conference on Computer Vision and Pattern Recognition (CVPR). IEEE, pp 4510–4520

49. Wu S, Zhong S, Liu Y (2018) Deep residual learning for image steganalysis. Multimed Tools Appl 77:10437–10453. https://doi.org/10.1007/s11042-017-4440-4

50. He K, Zhang X, Ren S, Sun J (2016) Deep Residual Learning for Image Recognition. In: IEEE Conference on Computer Vision and Pattern Recognition (CVPR). pp 770–778

51. Chollet F (2016) Xception: Deep Learning with Depthwise Separable Convolutions. In: IEEE Conference on Computer Vision and Pattern Recognition (CVPR). pp 1251–1258

52. Ioffe S, Szegedy C (2015) Batch normalization: Accelerating deep network training by reducing internal covariate shift. In: Proceedings of the 32nd International Conference on Machine Learning. PMLR, pp 448–456

53. Szegedy C, Ioffe S, Vanhoucke V, Alemi A (2016) Inception-v4, Inception-ResNet and the Impact of Residual Connections on Learning. In: Thirty-First AAAI Conference on Artificial Intelligence. pp 4278–4284

54. Zhang X, Zhou X, Lin M, Sun J (2017) ShuffleNet: An Extremely Efficient Convolutional Neural Network for Mobile Devices. In: IEEE Conference on Computer Vision and Pattern Recognition (CVPR). pp 6848–6856

55. Zoph B, Vasudevan V, Shlens J, Le Q V (2018) Learning transferable architectures for scalable image recognition. In: IEEE Conference on Computer Vision and Pattern Recognition (CVPR). pp 8697–8710

56. Jia Deng, Wei Dong, Socher R, Li-Jia Li, Kai Li, Li Fei-Fei (2009) ImageNet: A large-scale hierarchical image database. In: 2009 IEEE Conference on Computer Vision and Pattern Recognition. IEEE, pp 248–255

57. Pirnia P, Duhaime F, Manashti J (2018) Machine learning algorithms for applications in geotechnical engineering. In: Proceedings 71st Canadian Geotechnical Conference. p paper 339

58. Lang N, Irniger A, Rozniak A, Hunziker R, Wegner JD, Schindler K (2021) GRAINet: mapping grain size distributions in river beds from UAV images with convolutional neural networks. Hydrol Earth Syst Sci 25:2567–2597

59. McFall B, Young D, Fall K, Krafft D, Whitmeyer S, Melendez A, Buscombe D (2020) Technical feasibility of creating a beach grain size database with citizen scientists. Coast Hydraul Lab (US), Eng Res Dev Cent. https://doi.org/10.21079/11681/36456

60. Yu D, Eversole A, Seltzer M, et al (2014) An Introduction to Computational Networks and the Computational Network Toolkit (Invited Talk). In: 15th Annual Conference of the International Speech Communication Association

61. Goodfellow I, Bengio Y, Courville A (2016) Deep Learning. MIT press, Cambridge

62. Rajpura PS, Bojinov H, Hegde RS (2017) Object Detection Using Deep CNNs Trained on Synthetic Images. arXiv Prepr arXiv170606782

63. Frid-Adar M, Diamant I, Klang E, Amitai M, Goldberger J, Greenspan H (2018) GAN-based synthetic medical image augmentation for increased CNN performance in liver lesion classification. Neurocomputing 321:321–331. https://doi.org/https://doi.org/10.1016/j.neucom.2018.09.013

64. Duhaime F, Pirnia P, Manashti J, Temimi M, Toews M (2021) Particle size distribution from photographs : comparison of synthetic and real granular material images

65. Pirnia P, Duhaime F, Ethier Y, Dubé J-S (2019) ICY: An interface between COMSOL multiphysics and discrete element code YADE for the modelling of porous media. Comput Geosci 1

23:38–46. https://doi.org/10.1016/j.cageo.2018.11.002
25

66. Park S, Kwak N (2017) Analysis on the Dropout Effect in Convolutional Neural Networks. In: Lecture Notes in Computer Science (including subseries Lecture Notes in Artificial Intelligence and Lecture Notes in Bioinformatics). pp 189–204

67. Dahl GE, Sainath TN, Hinton GE (2013) Improving deep neural networks for LVCSR using rectified linear units and dropout. In: 2013 IEEE International Conference on Acoustics, Speech and Signal Processing. pp 8609–8613

68. Scherer D, Müller A, Behnke S (2010) Evaluation of pooling operations in convolutional architectures for object recognition. Lect Notes Comput Sci (including Subser Lect Notes Artif Intell Lect Notes Bioinformatics) 6354 LNCS:92–101. https://doi.org/10.1007/978-3-642-15825-4_10

69. Srivastava N, Geoffrey Hinton, Alex Krizhevsky, Ilya Sutskever, Ruslan Salakhutdinov (2014) Dropout: A Simple Way to Prevent Neural Networks from Overfitting. J Mach Learn Res 15:1929–1958

70. Karpathy A CS231n: Convolutional Neural Networks for Visual Recognition. http://cs231n.github.io/convolutional-networks/#layers

71. Simonyan K, Zisserman A (2014) Very Deep Convolutional Networks for Large-Scale Image Recognition. In: 3rd International conference on learning representations (ICLR). https://arxiv.org/abs/1409.1556, pp 1–14

72. Weiss K, Khoshgoftaar TM, Wang D (2016) A survey of transfer learning. J Big Data 3:9. https://doi.org/10.1186/s40537-016-0043-6

73. Russakovsky O, Deng J, Su H, et al (2015) ImageNet Large Scale Visual Recognition Challenge. Int J Comput Vis 115:211–252. https://doi.org/10.1007/s11263-015-0816-y

74. Nikolenko SI (2021) Synthetic data for deep learning. Springer, Cham

75. Sudhakar J, Adhikari GR, Gupta RN (2006) Comparison of Fragmentation Measurements by Photographic and Image Analysis Techniques. Rock Mech Rock Eng 39:159–168. https://doi.org/10.1007/s00603-005-0044-9